\documentclass[letterpaper, 10 pt, conference]{ieeeconf}  

\IEEEoverridecommandlockouts                              

\usepackage{hyperref}
\usepackage{graphicx}
\usepackage{xcolor}
\usepackage{balance}
\usepackage{amsmath,amssymb,amsfonts}
\usepackage[normalem]{ulem}
\definecolor{darkcyan}{rgb}{0.0, 0.55, 0.55}
\definecolor{mygreen}{rgb}{0.5, 0.9, 0.5}
\newcommand{\sectionref}[1]{Section~\ref{#1}}
\newcommand{\figureref}[1]{Figure~\ref{#1}}

\newcommand{\Equationref}[1]{(\ref{#1})}

\title{\LARGE \bf
Tunable Dynamic Walking via Soft Twisted Beam Vibration*
}

\author{Yuhao Jiang$^{1}$, Fuchen Chen$^{2}$, and Daniel M. Aukes$^{3}$
\thanks{*This work is supported by the National Science Foundation Grant No. 1935324. \textit{(Corresponding author: Daniel Aukes)}}
\thanks{$^{1}$School for Engineering of Matter, Transport and Energy, Fulton Schools of Engineering, Arizona State University, Tempe, AZ, 85281, USA
        {\tt\footnotesize yuhao92@asu.edu}}%
\thanks{$^{2}$The Polytechnic School, Fulton Schools of Engineering, Arizona State University, Mesa, AZ, 85212, USA {\tt\small fchen65@asu.edu}}        
\thanks{$^{3}$The Polytechnic School, Fulton Schools of Engineering, Arizona State University, Mesa, AZ, 85212, USA
        {\tt\footnotesize danaukes@asu.edu}}%
}

\begin{document}
\maketitle

\begin{abstract}
We propose a novel mechanism that propagates vibration through soft twisted beams, taking advantage of dynamically-coupled anisotropic stiffness to simplify the actuation of walking robots. Using dynamic simulation and experimental approaches, we show that the coupled stiffness of twisted beams with terrain contact can be controlled to generate a variety of complex trajectories by changing the frequency of the input signal. This work reveals how ground contact influences the system's dynamic behavior, supporting the design of walking robots inspired by this phenomenon. We also show that the proposed twisted beam produces a tunable walking gait from a single vibrational input.
\end{abstract}

\section{Introduction}
\label{intro}

Actuation and its transmission through soft robotic systems have driven extensive study in recent decades~\cite{soft_actuator_review,act9010003,Fitzgerald2020}. Unlike actuation in traditional rigid-body robotic systems -- which relies on motors, gears, shafts, and belts to actuate and transmit power -- the morphology of soft actuators can be deformed to subsequently alter body shapes and drive robots by stimulating or deforming soft materials. While numerous soft actuators have been developed to drive  soft robots in applications like human-robot interaction, bio-inspired robots, and wearable robotic systems, the power of these systems is usually low and  actuators are usually bulky. Moreover, due to the non-linearity of hyper-elastic materials and the complexity of powered soft systems,  dynamic modeling is challenging and thus can be under-utilized during the design process. 

In this paper, we propose a novel actuating method for walking robots using the coupled compliance of soft twisted beams with ground contact. This mechanism transforms simple, periodic input motion into complex cyclic motions when contact is made with the ground. More specifically, in this paper, we show how this phenomenon can be adopted to generate tunable forward and backward walking by controlling the input frequency. 
This study fits under the umbrella of a new class of devices we call "Soft, Curved, Reconfigurable, Anisotropic Mechanisms" (SCRAMs), which we have previously studied in the context of pinched tubes\cite{9341109,9479208,9479201}, and buckling beams\cite{9244584,9369911}. By taking advantage of the shape and material properties in soft structures, complex actuation signals for generating complex motion can be consolidated and simplified.

\begin{figure}[!tp]
    \centering
    \includegraphics [width=0.49\textwidth]{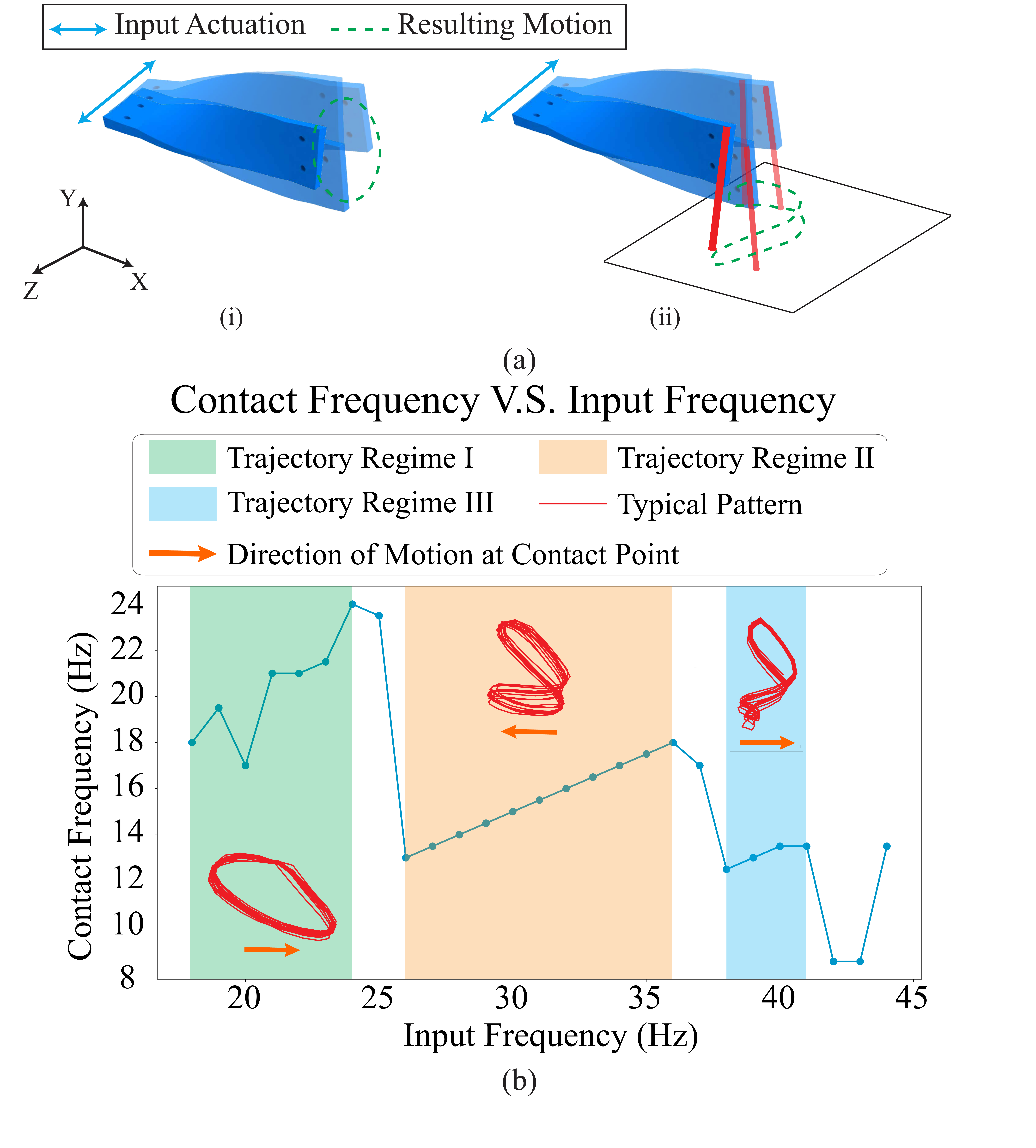}
    \caption{\textbf{Concept demonstration and beam design.} 
    \textbf{(a):} Conceptual demonstration of the operation principle: (i) without terrain contact; (ii) with terrain contact.
    \textbf{(b):} Single beam contact test result: contact frequency as a function of input frequency.
    }
    \label{fig:concept}
\end{figure}

\figureref{fig:concept}(a) demonstrates the proposed vibration propagation concept. In (i), a soft, twisted beam under a linear vibratory input (as shown by the blue arrow) generates a repeating, semicurcular trajectory at the tip, as shown by the dashed green path. 
This motion, with terrain contact, as shown in (ii), results in a more complex motion that can be further adapted for robot walking. In this paper, we show that the contact frequency, direction of motion at the contact point, as well as the resulting motion path can be controlled by the input frequency as shown in \figureref{fig:concept}(b).

The rest of the paper is organized as follows.  \sectionref{sec:background} discusses related prior work, while \sectionref{sec:contribution} summarizes the contributions of this paper. In \sectionref{sec:design}, we describe the design and manufacturing of  prototype beams. \sectionref{sec:system_modeling} presents the FEA simulation of the soft twisted beam free vibration and the soft twisted beam vibration with terrain contact using an analytical model. \sectionref{sec:labtest} subsequently discusses the experimental validation of our concept.
A walking robot prototype is then presented in \sectionref{sec:walk}. The test results and limitations are discussed in \sectionref{sec:discussion}; the paper then concludes in \sectionref{sec:conclusion} along with a discussion of planned future work.
\subsection{Background}
\label{sec:background}

Helical shapes, twisted surfaces, and chirality are found throughout the natural world, such as DNA molecule~\cite{Watson1974}, Erodium cicutarium seeds~\cite{Evangelista2011}, Bauhinia seed pods~\cite{doi:10.1126/science.1203874}, human sperm~\cite{doi:10.1126/sciadv.aba5168}, snails~\cite{Schilthuizen2005}, and cucumbers\cite{doi:10.1126/science.1223304}. These natural phenomena have aroused a series of theoretical studies regarding the self-assembly and transition of complex helical strands such as cables, ropes, and ribbons~\cite{10.1115/1.3101684,PhysRevLett.93.158103,Ghafouri2005}. 

Static and dynamic models essential for understanding rigid, pre-twisted beams were originally developed in the context of applications such as helicopter rotor blades, wind turbines, fans, and turbo-engines. \cite{Hodges1990,Hodges2009}. Recent studies have focused on wave propagation in twisted beams, for providing insights on the motion of waveguides~\cite{MUSTAPHA201246}. These studies, while applied to nano-scale systems at high frequencies, reveal how waves are altered as they are transmitted through twisted, continuum systems.

Inspired by nature and the mathematical properties of anisotropic, curved, chiral, and helical shapes, scientists have also developed soft systems that can generate complex asymmetric motion for use in actuation ~\cite{Wang2016,Sun2017,Cheng2021} and sensing\cite{Torsi2008, Qaiser2021}. Various soft actuation methods have been proposed to utilize the stiffness and the geometry change of continuous curved surfaces for locomotion~\cite{Yang2021,Zhai2020,9341109,9479208}. Twisting mechanisms have also been applied in the actuation of robotic fingers~\cite{6248225} and twisting tube actuators~\cite{Li2021}. Zhao et al.~\cite{Zhao2022} have 
developed a twisting ribbon robot that can roll and maneuver in unstructured environments. The above works demonstrate that curved geometry can play an important role in establishing tuned dynamic complex gaits in soft or flexible robotic systems. Maruo et al.~\cite{Maruo2022} have proposed a similar mechanism that uses structural anisotropy and cyclic vibrations to create complex motions for manipulation. Differing from~\cite{Maruo2022} where the mechanism was used for object manipulation, this paper studies the capability of anisotropic soft twisted beams interacting with the ground to generate complex walking locomotions via periodic actuation input.

Cyclic vibration has been adopted in the prior art as a power source in terrestrial locomotion.  Bristlebots -- a well-known and simple class of walking mechanisms -- utilize vibration-based actuation and inclined, oriented bristles to move forward; studies have been conducted on various scales of bristlebots~\cite{Gandra2019,Kim2020,10.1115/1.2899031}. Li et al have developed an insect-scale robot actuated by stretchable dielectric elastomers to achieve ratcheting walking locomotion~\cite{Li2019}. While the above research demonstrates the capacity for vibration-based actuation to drive terrestrial robots, the type of motions observed in these systems is limited due to the direct connection to the input actuator. This has artificially limited applications to  simpler tasks on lower-complexity terrains. In contrast, we propose mechanisms for establishing more complex leg dynamics using soft and compliant twisted beams in this paper, which can be tuned via the geometric, inertial, and material parameters of our design and used to simplify the control signals typically associated with multi-DOF walking robots.

\subsection{Contributions}
\label{sec:contribution}

The contributions of this paper may be summarized as follows: 
1) A new mechanism has been proposed for generating walking locomotion using soft twisted beams under interaction with the ground; 2) A model has been developed to describe the dynamic behavior of the highly nonlinear soft twisted beam using a pseudo-rigid body model for fast simulation.
3) Using simulations and experimental platforms with minimal constraints, we have demonstrated how walking direction and speed can be tuned by the frequency of the input actuator. 
\section{System Modeling}
\label{sec:system_modeling}
This section describes FEA and analytical modeling approaches for understanding the dynamic behavior of the proposed  soft twisted beam vibration as well as the resulting motion when interacting with the terrain surface. The FEA simulation results demonstrate the resulting motion of the freely vibrating soft twisted beam as a function of twist angle and driving frequency. The pseudo-rigid-body model demonstrates the walking locomotion of the vibrating soft twisted beam with terrain contact as a function of driving frequency.

\subsection{Dynamic modeling using FEA approach}
\label{sec:simulation}

We conducted a series of dynamic simulations using FEA model in PyChrono~\cite{chrono}, The results of the simulations demonstrate how input frequency, beam chirality, and the magnitude of beam twist angle $\phi$ as shown in \figureref{fig:lab_test_setup}(a) alter the dynamic motion of the beam.


\subsubsection{FEA model setup}
\label{sec:FEA_setup}
\-

We developed the FEM-based dynamic model seen in \figureref{fig:simulation}(a).  It consists of a 120-element mesh generated from a single layer of 6-field Reissner-Mindlin shells. The mesh geometry replicates the beam design outlined in \sectionref{sec:design} and the material properties for TPU came from its datasheet.

The input actuator  shakes the proximal end of the beam along the z-axis as shown in \figureref{fig:simulation}(a). The input signal is represented by
\begin{equation}
\label{eqa:input}
x = A\sin(2\pi ft),
\end{equation}
where $x$ is the actuation travel position with the unit of mm, $f$ is the rotating frequency of the motor in Hz and $A$ is the amplitude in mm with $A=2$\,mm.

\subsubsection{Input frequency V.S. resulting motion}
\label{sec:FEA_frequency}
\-

The coupled stiffness of twisted beams can be exploited by exciting it at specific frequencies to create highly differentiated motion.
To demonstrate this effect, we swept the input frequency from  $f=1$\,Hz to $f=45$\,Hz in 1\,Hz increments. The trajectory of the beam's distal end was recorded throughout the simulation and is shown in \figureref{fig:simulation}(f). As can be seen, the beam's trajectory varies significantly in shape and size as a function of input frequency. At certain input frequencies such as 9\,Hz, 17\,Hz, 25\,Hz, the trajectory exhibits an oval-like shape, whereas at frequencies such as 1\,Hz and 41\,Hz the trajectory appears more linear.


\subsubsection{Beam twist V.S. resulting trajectory}
\label{sec:fea_twist}
\-

\begin{figure*}[!tp]
    \centering
    \vspace*{7pt}
    \includegraphics[width=1\textwidth]{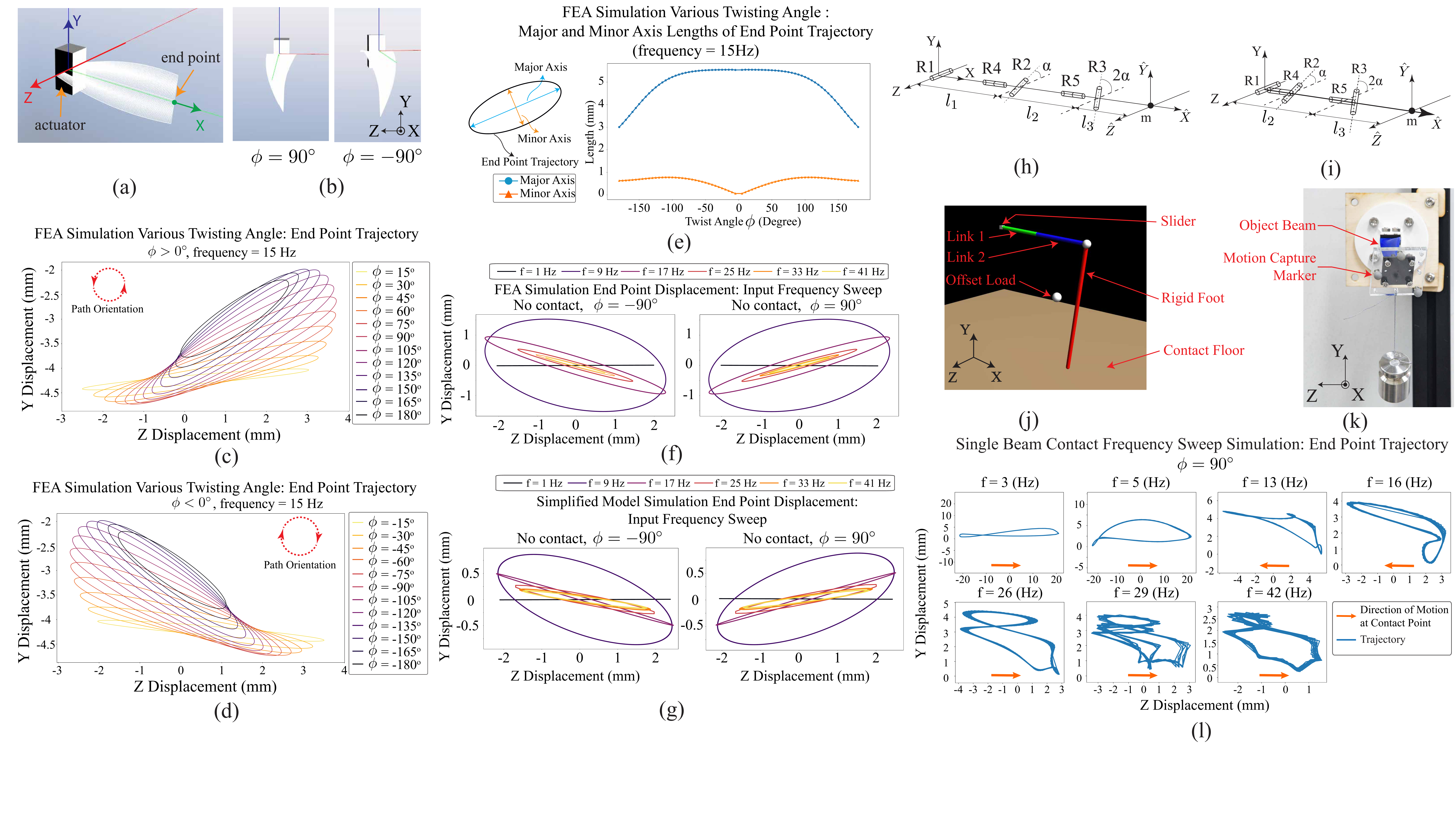}
    \caption{\textbf{Results from FEA simulations.} 
    \textbf{(a:)} Beam FEA mesh model.
    \textbf{(b:)} Beam with right-handed and left-handed chirality.
    \textbf{(c):} Beam end point trajectory results from FEA simulation with positive twist angle $\phi$.
    \textbf{(d):} Beam end point trajectory results from FEA simulation with negative twist angle $\phi$.
    \textbf{(e):} Major and minor axis lengths of beam end point trajectory with respect to twist angle $\phi$.
    \textbf{(f):} Beam end point trajectory results from FEA frequency sweep simulation.
    \textbf{(g):} Beam end point trajectory results from frequency sweep simulation using the proposed analytical model.
    \textbf{(h):} The proposed analytical model.
    \textbf{(i):} The simplified analytical model.
    \textbf{(j):} The contact simulation setup in MuJoCo.
    \textbf{(k):} The lab test setup for beam dynamic characterization.
    \textbf{(l):} Beam end point trajectory results from frequency sweep simulation with contact using the proposed analytical model.
    }
    
    \label{fig:simulation}
\end{figure*}

A beam's magnitude of twist plays an important role in the generation of elliptical motion, while its chirality (twist direction) can be used to mirror the patterns observed at different magnitudes.
We explored the relationship between beam twist angle $\phi$ and its resulting trajectory through a pair of studies. In the first study, we modeled a series of beams with identical dimensions but a range of twist angles from $\phi=0^\circ$ to $\phi=180^\circ$ with a step of $5^\circ$. The input amplitude and frequency was held constant at $f=15$\,Hz and $A=2$\,mm. The distal end's trajectory was recorded during the simulation; the selected result is shown in \figureref{fig:simulation}(c).  As the twist angle $\phi$ increases, the output trajectory's orthogonal motion (along the $Y$ axis) grows. To better understand the nature of the shapes generated, we approximated  each trajectory as an elliptical path, identified the major and minor axes of the approximate ellipses at each frequency, and then measured their length.  The results, shown in \figureref{fig:simulation}(e), highlight how twist magnitude and the resulting coupling of stiffness play a role in the evolution of elliptical paths in twisted beams. Based on this result, the twist angle $\phi$ of the prototype beams is set as $\phi=90^\circ$ and $\phi=-90^\circ$ for the more distinguished spans in both major and minor axis.


In the second set of simulations, we compared beams of equal magnitude but opposite direction $(\phi_1 = -\phi_2)$, as shown in  \figureref{fig:simulation}(b).  As can be seen in \figureref{fig:simulation}(c) and (d), beams of equal magnitude but opposite chirality result in trajectories mirrored over the Y-axis (the beams' axis of symmetry).  It should be noted that not only is the elliptical shape mirrored, but the path orientation along that shape is inverted or mirrored as well.  This is highlighted in \figureref{fig:simulation}(c) and (d) by the red dashed arrows. 


\subsection{Pseudo-rigid-body modeling}
\label{sec:approx_model}

In order to fast simulate and explore the beam dynamic behavior with contact, a simplified model that is less computationally expensive is strongly demanded. Thus, we employed the pseudo-rigid-body model to describe the dynamic behavior of twisted beams over time. This approach is described below.
\subsubsection{Pseudo-rigid-body model setup}
\-

Off-diagonal coupling parameters, along with hyper-elastic material models, makes the dynamics of twisted soft systems more complex than classical approaches such as Euler–Bernoulli models can approximate.  Fundamental research has analyzed the behavior of pre-twisted beams. Hodges~\cite{Hodges1990} presents a variational formulation for the dynamics of stiff, pre-twisted beams, and in a follow-up paper proposes a geometrically intrinsic dynamic model of twisted beams~\cite{Hodges2009}.  These approaches use a geometric and kinematic approach to build the dynamic relations between beam loading and deformation. Banerjee~\cite{Banerjee2001} presents a free vibration analysis of twisted beams, developing a dynamic stiffness matrix that describes the force-displacement relationship at the nodes of a harmonically-vibrating twisted cantilever beam with end loading. This work highlights the flexural, coupled displacements  across two orthogonal planes, demonstrating the potential for approximate representations of twisted beam dynamics using simplified models with two cooperative linear motions in two mutual-orthogonal planes. 

We propose a pseudo-rigid-body model with revolute springs attached to a number of joints subdividing the beam. We have used a linear spring-damper model of the form 
\begin{equation}
\label{eq:joint}
\tau = k\theta+b\dot{\theta}
\end{equation}
to describe the moments about each joint, where $\tau$ represents the torque about each joint, $k$ represents the linear spring constant in bending, $b$ represents linear joint damping, and $\theta, \dot{\theta}$ represent the local rotation and rotational velocity, respectively, of each joint from its unloaded, natural shape.  Since the cross-sectional area of each beam is constant along its axial length, the spring stiffness constant $k$ represents a distributed bending stiffness about three revolute joints --  R1, R2, and R3 -- which are distributed perpendicularly along the beam's axial direction, as seen in the complete model in Figure.~\ref{fig:simulation}(h). Two additional revolute joints -- R4, and R5 -- are aligned with the beam's local axial direction and capture the twist of the beam, represented by $\phi$. The same spring-damper model as \Equationref{eq:joint} was applied to represent the twisting stiffness on these two joints.

Together, these joints exhibit the same coupled stiffness of twisted beams observed in experiments, as demonstrated through our FEA simulation performed later in this section.  Based on the results from \cite{Howell1992, Howell1996}, the location of joints in a compliant, cantilever-style pseudo-rigid-body model under large-deflections should not be evenly distributed along the beam; we thus parameterize $l_1$, $l_2$, $l_3$ as the distances between R1-R2, R2-R3, and R3 - distal end, respectively. The total length of the beam, $l = l_1+l_2+l_3$,  is set to be identical to the prototype as $l = 50mm$.

The mass is evenly distributed by the density of TPU $\rho = 1210 kg/m^3$. The total mass of the beam, or the sum of all links' mass, is equal to the prototypes' mass of $m = 5.17g$. The mass of each link is proportional to its link length, with $m_k = \frac{m}{l}\cdot l_k$, where $k = 1,2,3$.

\subsubsection{Model fitting}
\-

A set of dynamic experiments was conducted to obtain the motion of the end of the beam when released from an initial deformed state. The test setup can be seen in \figureref{fig:simulation}(k). At the beginning of the test, the beam was deformed with a 200\,g load applied to the end. The load was instantaneously released from the beam while the position of the beam's tip was recorded as the beam returned to rest at its natural unloaded position. Three optical tracking markers were attached to the end of the beam to obtain the tip's motion.  After the data was recorded, a differential evolution optimizer\cite{Storn1997} was then implemented to fit the model variables ($k$, $b$, $l_1$, $l_2$, $l_3$) by minimizing the averaged error between simulation marker position data ($M_i$) and the reference data from experiments ($\hat{M_i}$). The objective function is shown below:

\begin{equation}
Min\Bigg\{ \sqrt{\sum_{j=0}^{n} \sum_{i=1}^{3} \left[(M_i(j) - \hat{M_i}(j))^2 \right] /(3n) } \Bigg\}
\end{equation}

The optimization variable set is defined by ($k$, $b$, $l_1$, $l_2$, $l_3$), where $l_3 = 50 - l_1 - l_2$.  In this fitting progress, the proposed model was simulated in MuJoCo~\cite{mujoco} and Python. We observed that $l_1$ tended to converge at the minimum bound of $1$\,mm; We therefore simplified the model by setting $l_1 = 0$, which yields the variable set as ($k$, $b$, $l_2$, $l_3$), where  $l_3 = 50 - l_2$. The optimizer finally converged with a mean absolute error of 3.49\,mm, where $k=0.340$\,N/rad, $b = 0.0029$, $l_2 = 23.66$\,mm, $l_3 = 26.34$ \,mm. 

We conducted the same simulation as in~\sectionref{sec:FEA_frequency} to show that the fitted simplified model delivers similar output to the FEA simulation. During the simulation, we commanded the input motor to oscillate and actuate one side of the beam, as described in~\sectionref{sec:FEA_frequency} from 1\,Hz to 45\,Hz, while the endpoint displacement on the other side of the beam was recorded. The resulting trajectory is shown in \figureref{fig:simulation}(g). As the input frequency increases, the endpoint motion shows similar motions to the FEA simulation, which transits from a line to an oval-like orbit that begins to tilt at higher frequencies. The averaged time cost for a 10\,s simulation with an Intel i7-7900K CPU and 32GB RAM was drastically shortened from 82.5\,s using FEA model to 1.2\,s using the newly proposed simplified model.

 \subsubsection{Simulation of single beam vibration with contact}
\-

Using the newly proposed pseudo-rigid-body model, we conducted a series of beam vibration simulations with contact in MuJoCo. The test setup,  as shown in \figureref{fig:simulation}(j), is identical to that described in \sectionref{sec:contact}. During the simulation, the slider is actuated to sweep from $f=1$\,Hz to $f=45$\,Hz using \Equationref{eqa:input} with amplitude $A=2$\,mm while the beam's end point position is recorded. The resulting trajectory and the direction of motion at the contact point are shown in~\figureref{fig:simulation}(l). As can be seen, the resulting motion differs from the free vibrating beam due to contact with the floor. A figure '8' loop is observed at the input frequency $f=16$\,Hz and $f=26$\,Hz. Moreover, the direction of motion at the contact point, as indicated by orange arrows, also alters as a function of the input frequency. 

\section{Design and manufacturing of the prototype beam}
\label{sec:design}

We designed and manufactured a series of prototypes to validate the proposed concept. 3D printing was selected to reduce manufacturing time and to permit a broad design space. Because hard printable plastics must be printed with very thin geometries and at higher precision to achieve the desired range of leg stiffnesses, in order to maintain a wide design space, we selected soft  printable materials that could be printed at millimeter to centimeter scales, more than 30 layers thick, while achieving the desired range of leg stiffness in all dimensions.
We compared two commercial soft filaments: thermoplastic elastomer (TPE)\footnote{Arkema 3DXFLEX™ TPE}, with a Shore hardness of 92A, and thermoplastic polyurethane (TPU)\footnote{Ultimaker TPU 95A}, with a Shore hardness of 95A. The Young's modulus of the TPE selected is reported as 7.8\,MPa in the datasheet, whereas the Young's modulus of the TPU is reported as 26\,MPa. Although the difference in the hardness between the two materials is relatively small, the TPU 95A 's higher stiffness supports our target payload and deflects less at the same dimensions compared to the TPE, while demonstrating the dynamic behavior desired for terrestrial locomotion. Thus, we selected the TPU 95A as the prototyping material.

Based on the simulation results in \sectionref{sec:fea_twist}, a number of prototypes with $\phi=90^\circ$ and $\phi=-90^\circ$ at the same length($l$), width($w$), and thickness($t$) were manufactured, as shown in~\figureref{fig:lab_test_setup}(b), the beam is right-handed chiral if $\phi > 0$ and left-handed if $\phi < 0$. Design diagram is shown in~\figureref{fig:lab_test_setup}(a), further design parameters can be found in Table~\ref{tab:design}.



\begin{table}[h]
\begin{center}
\caption{Design Parameters}
    \begin{tabular}{ | c | c | c | c | p{5cm} |}
    \hline
    Parameter & Symbol & Value & Unit\\
    \hline
    Beam length & $l$ & 50 & mm\\
    Beam width & $w$ & 20 & mm\\
    Beam thickness & $t$ & 3 & mm\\
    Beam total twist angle & $\phi$ & 90 & degree\\
    Beam segmental twist angle & $\alpha$ & 45 & degree\\
    \hline
    \end{tabular}
    \label{tab:design}
\end{center}
\end{table}

\section{Prototype Tests}
\label{sec:labtest}

The results of our experiments demonstrate how vibrating, twisted beams with terrain interactions exhibit similar behavior in real life to model-based results. 
\subsection{Single Beam Contact Test}
\label{sec:contact}

This experiment demonstrates how the output trajectory and its orientation can be influenced by the input signal driving frequency in the presence of highly nonlinear ground interactions. This section demonstrates a relatively constrained, prescribed experiment, whereas the next section demonstrates the same phenomonon observed in a less prescribed manner with a free-walking platform.

\begin{figure}[!tp]
    \centering
    \vspace*{3pt}
    \includegraphics[width = 0.49\textwidth]{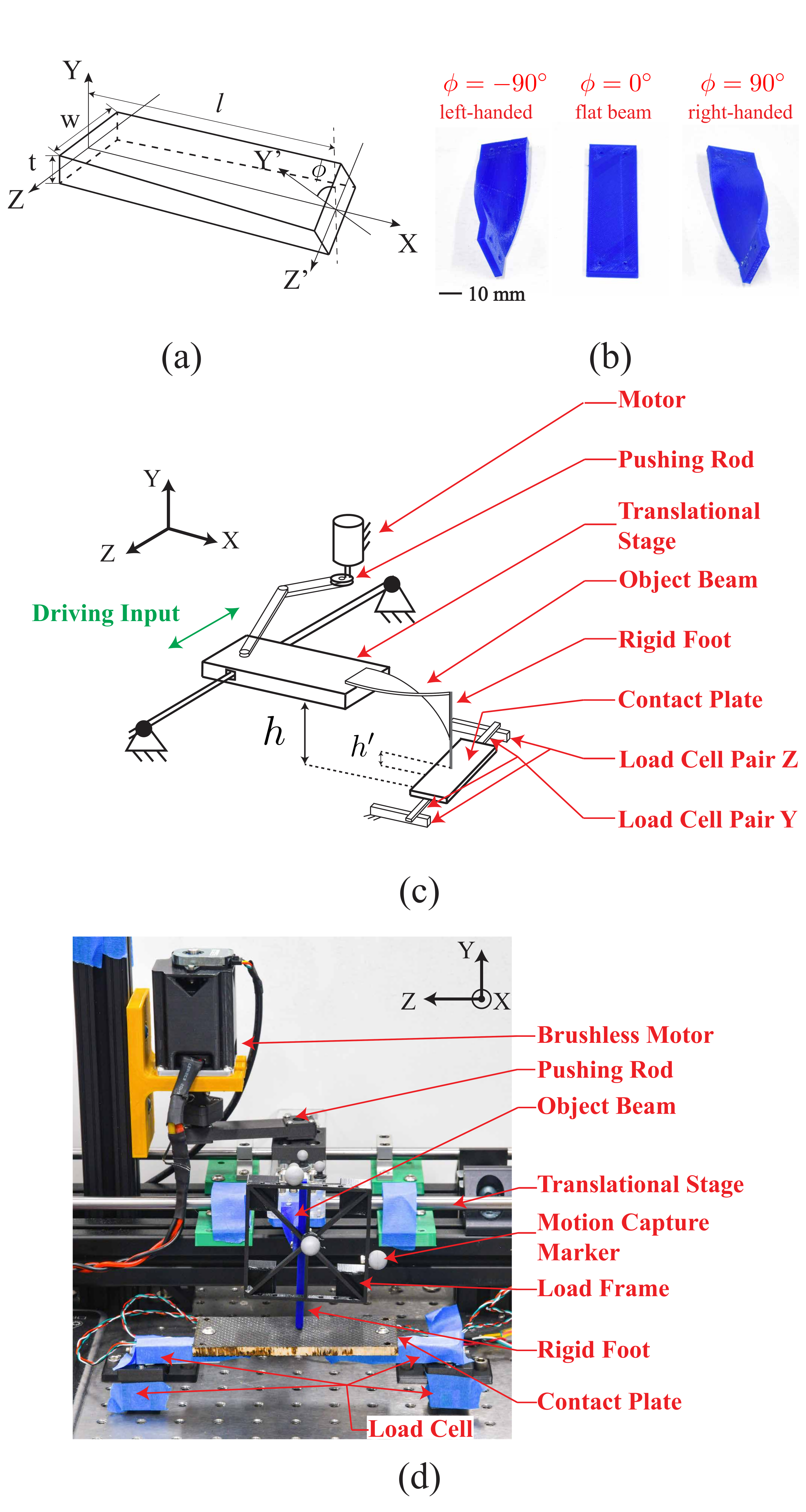}
    \caption{\textbf{Lab test setups.}
    \textbf{(a):} Design diagram of the twisted beam.
    \textbf{(b):} Beam prototype samples.
    \textbf{(c):} Sketch of lab test setups.
    \textbf{(d):} Lab test setup for single beam contact tests
    }
    \label{fig:lab_test_setup}
\end{figure}

\begin{figure}[tp]
    \centering
    \vspace*{4pt}
    \includegraphics[width=0.48\textwidth]{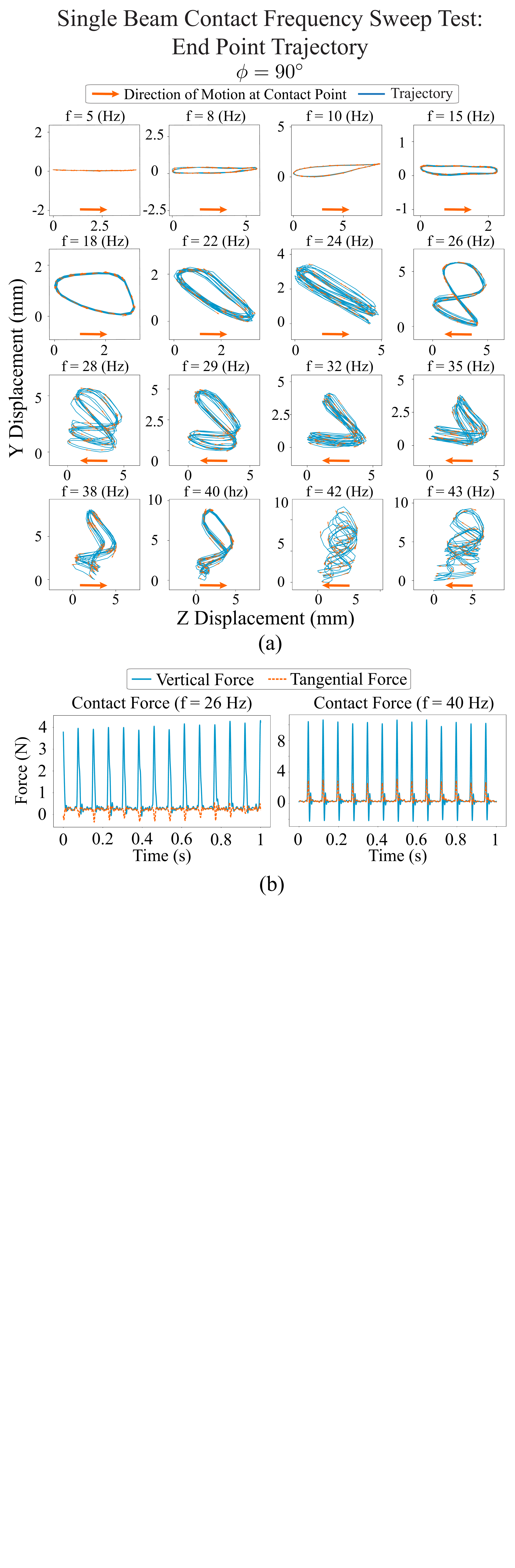}
    \caption{\textbf{Results from single beam contact tests.}
    \textbf{(a):} Selected end point trajectories.
    \textbf{(b):} Contact force data in two directions.
    }
    \label{fig:lab_contact}
\end{figure}

The test setup in \figureref{fig:lab_test_setup}(c) and (d) shows a linear stage whose oscillating, forward-backward motion is dictated by the rotating crank of a brushless motor\footnote{ODrive Dual Shaft Motor D6374 - 150KV}.
The motor is controlled by an ODrive\footnote{Odrive V3.6 High Performance Motor Control.} motor control board. 
We again use \Equationref{eqa:input} to control the speed of the motor, with $A=2$\,mm, and $f=\{1-40\}$\,Hz.
The beam is mounted to the linear stage and optical tracking markers are mounted to the proximal and distal ends of the beam.  An OptiTrack Prime 17W optical motion tracking system is then used to track the position of the system at a rate of 360\,Hz.
A plate with four load cells mounted perpendicularly in sets of two, to measure contact forces between the leg and ground along the Y and Z axes, as shown in \figureref{fig:lab_test_setup}(c) (normal and tangential to the ground, respectively). The test setup is shown in \figureref{fig:lab_test_setup}(d) and the test results are shown in \figureref{fig:lab_contact}. The beam sample with $\phi = 90^\circ$ was used, and the mass of the foot is represented by a 20\,g load attached to the lower left corner of the load frame. The length of the rigid foot is $66.5$\,mm, and the distance between the translational stage and the plate is $h= 72$\,mm as shown in \figureref{fig:lab_test_setup}(c). Therefore the contact distance between the foot at its unload, natural position and the plate, as depicted by $h'$ in \figureref{fig:lab_test_setup}(c) is fixed at $5.5$\,mm.

Typical trajectories have been selected and plotted in \figureref{fig:lab_contact}(a). As can be seen, the trajectory evolves as a function of input frequency. In the low-frequency region, where the input frequency is less than 18\,Hz, contact interactions dominate the motion observed in the leg, because the "foot" never breaks contact with the ground.  This results in trajectories which are a flat line along the Z axis. As the input frequency increases to 26\,Hz, ground contact becomes more intermittent and the leg's motion becomes dominated by its own dynamic properties.  This results in trajectories that look like a figure '8', or a loop with a single inversion.  At the point of contact, the inverted trajectory results in a change in the direction of motion, shown by the orange arrows in \figureref{fig:lab_contact}(a). At frequencies higher than ~38\,Hz, the trajectory inverts a second time and the direction of motion at the point of contact reverses again. 

The tangential forces measured by the load cells also capture direction changes at the same transition frequencies. In \figureref{fig:lab_contact}(b), two typical force data are plotted at frequencies of 26\,Hz and 40\,Hz. By comparing the tangential forces, one can see that the direction is opposite, in line with the change in motion observed in \figureref{fig:lab_contact}(a). The vertical force data can be used to capture the contact frequency, which is not necessarily the same as the driving frequency. Since contact dominates at frequencies below 18\,Hz, we focus on frequencies from 18\,Hz to 44\,Hz. The result is shown in \figureref{fig:concept}(b). We highlighted three distinct shapes observed using different colors. In each regime, the contact frequency increases with the input frequency.  At the transition frequencies noted previously (26\,Hz and 38\,Hz), the contact frequency drops by ($\frac{1}{2}$ and $\frac{1}{3}$, respectively), the same frequencies at which the foot's trajectory inverts itself and then reverses its direction of motion (and force) on the ground.



It should be noted that this experiment was conducted at a fixed height off the ground.  The next section explores how a less-constrained system exhibits similar behavior to produce controllable, walking gaits.

\subsection{Walking Tests}
\label{sec:walk}

This test demonstrates how the proposed twisted beam can be leveraged to produce a controllable walking gait that can be easily tuned from a single vibrational input.

\begin{figure}[!tp]
    \centering
    \vspace*{6pt}
    \includegraphics[width=0.48\textwidth]{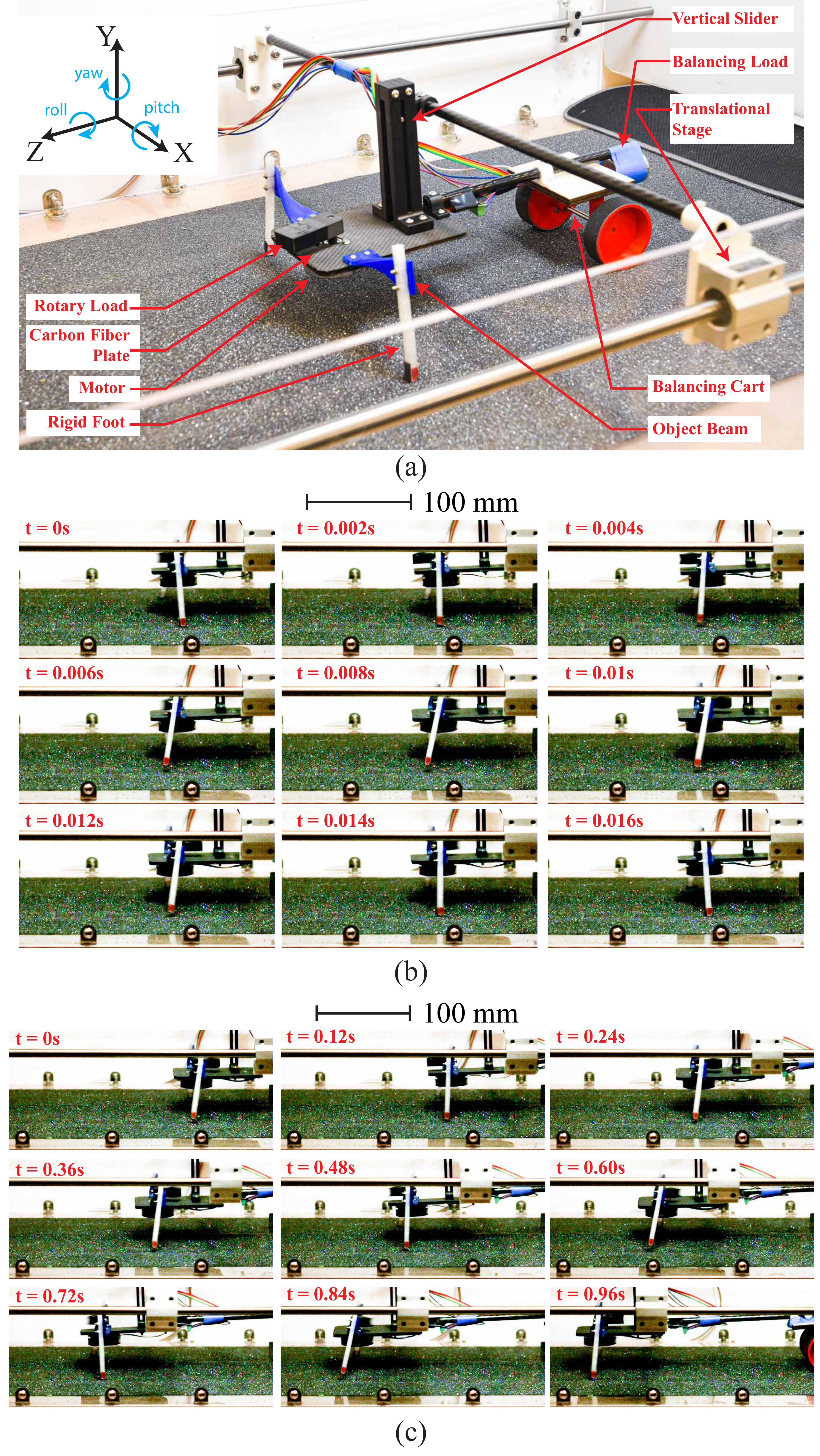}
    \caption{\textbf{Results from walking tests.}
    \textbf{(a):} Walking test setups.
    \textbf{(b):} One cycle of walking gait when walking forward.
    \textbf{(c):} Walking trajectory in 1\,sec.
    }
    \label{fig:walking}
\end{figure}

Two twisted beams serve as robot legs with $\phi=90^\circ$ and $\phi=-90^\circ$, respectively, are mounted in a mirrored fashion across the robot's sagittal plane to a carbon fiber plate. A Maxon brushless motor\footnote{Maxon EC 45 flat Ø42.9 mm, brushless, 30 Watt, with Hall sensors} along with a 40\,g offset load is fixed to the plate, serving as a rotary actuation input. The test setup is shown in \figureref{fig:walking}(a). A vertical slider  connects the robot to two translational stages so that the motion of the robot  is constrained along the x-axis and about the yaw axis, while the motion about and along the roll, pitch, z-axis, and y-axis is permitted.  A cart with a 100\,g load is attached to the robot's tail for support and balance. The total length of the walking platform is 295\,mm.

During this test, the motor was commanded to drive the robot at various frequencies from 1\,Hz to 80\,Hz in 1\,Hz increments. A high-speed camera \footnote{Edgertronic SC1, \href{https://www.edgertronic.com/our-cameras/sc1}{https://www.edgertronic.com/our-cameras/sc1}} was used to record the position of the robot at the rate of 1000\,fps. Test videos can be found in the supplemental video. \figureref{fig:walking}(b) presents a cycle of the walking gait at the actuating frequency of 65\,Hz. \figureref{fig:walking}(c) shows the trajectory of the robot in 1 second. In this test, the robot reached the averaged walking speed of 156.3\,mm/s with a 65\,Hz actuating input frequency. In addition to walking forward, the robot was also able to move backward at a speed of 35.7\,mm/s at an input frequency of 23\,Hz. This result demonstrates how foot motion can be tuned by altering the one-DoF actuation input frequency and shows great potential for controlling the walking direction and speed by tuning the input actuation frequency. 

\section{Discussion}
\label{sec:discussion}

Our results demonstrate, through simulation and single beam experiments with contact, that the coupled stiffness of twisted beams can be easily controlled to generate a variety of complex motions by simply changing the beam's input frequency.  The results presented above suggest a rich space for control, even from simple actuation sources. These experiments also reveal how highly nonlinear ground contact influences the system's dynamic behavior, which supports the design of walking robots inspired by this phenomenon. 

Our experiments progressively move from single beam contact tests to less-constrained studies of system motion with multiple legs in contact with the ground.  Through the successive release of constraints, we have demonstrated that the underlying dynamics continue to be influenced  by both beam design parameters and input signals.  As we continue to release constraints and add legs, we anticipate further challenges with regard to the synchronization of multi-legged systems against the complexity of multiple points of contact vibrating at high speed against the ground.  We believe that these topics are outside the scope of the current paper, in which we have primarily focused on the role of design and actuation inputs on single-beam behavior.

Some limitations have also been observed throughout the study. To begin with, 
we observed that the beam heats up over the course of a long data collection run, which alters material properties such as stiffness and elasticity, impacting results. 
To address this issue, future designs will integrate materials with lower viscoelastic loss modulus, higher temperature coefficient of Young's modulus, and optimized geometries to reduce shear stresses under vibration, in order to reduce the impact temperature plays on the system's shifting dynamic properties.
Another current limitation of this work is the lack of a full-body simulation of a multi-legged robot. Simulating our system is challenging because it involves multi-point, soft-body contact with the ground -- highly nonlinear interactions that require heavy computation.  
We plan to employ the newly proposed simplified beam model to simulate the system-level dynamics at faster rates.  Once developed, this simulation would permit mechanical design optimization and controller design for understanding the full suite of capabilities in this new legged robot.
\section{Conclusion}
\label{sec:conclusion}

In this paper, a mechanism for propagating  vibration through soft twisted beams with ground contact is proposed for simplifying the actuation of walking robots by taking advantage of these beams' dynamically-coupled anisotropic stiffness. A simplified model has also been proposed to fast simulate the nonlinear dynamic behavior of the soft twisted beam. Using dynamic simulation and experimental approaches, we have shown that the coupled stiffness of twisted beams with terrain contact can be controlled to generate a variety of complex trajectories by changing the
frequency of the input signal. This work also reveals how highly nonlinear ground contact influences the system's dynamic behavior, supporting the design of walking robots inspired by this phenomenon. 
Future work will explore manufacturing and design strategies for improving consistency between SCRAM elements and minimizing energy loss due to heat. Future work will also include studies on extending the versatility of this concept for locomotion in various media like water and air with gaits like swimming and flapping.






\section*{ACKNOWLEDGMENT}

This work is supported by the National Science Foundation Grant No.\,1935324.


\newpage

\balance

\bibliographystyle{ieeetr}

\end{document}